\documentclass[10pt, a4paper]{article}
\usepackage{lrec2022} 
\usepackage{multibib}
\newcites{languageresource}{Language Resources}
\usepackage{graphicx}
\usepackage{soul}
\usepackage{titlesec}
\titleformat{\section}{\normalfont\large\bfseries\center}{\thesection.}{1em}{}
\titleformat{\subsection}{\normalfont\SmallTitleFont\bfseries\raggedright}{\thesubsection.}{1em}{}
\titleformat{\subsubsection}{\normalfont\normalsize\bfseries\raggedright}{\thesubsubsection.}{1em}{}
\renewcommand\thesection{\arabic{section}}
\renewcommand\thesubsection{\thesection.\arabic{subsection}}
\renewcommand\thesubsubsection{\thesubsection.\arabic{subsubsection}}

\usepackage{epstopdf}
\usepackage[utf8]{inputenc}
\usepackage{multirow}
\usepackage{hyperref}
\usepackage{xstring}

\usepackage{multicol}

\usepackage{arydshln}
\usepackage{color}

\usepackage{xcolor}
\usepackage{colortbl}

\usepackage{enumitem} 

 \title{Tweet Emotion Dynamics:\\ Emotion Word Usage in Tweets from US and Canada}

\name{Krishnapriya Vishnubhotla$^{\ast}$$^{\ddagger}$ and Saif M. Mohammad$^{\dagger}$}

\address{$^{\ast}$Department of Computer Science, University of Toronto \\ 
        $^{\ddagger}$Vector Institute for Artificial Intelligence \\
                 \texttt{vkpriya@cs.toronto.edu} \\
         $^{\dagger}$National Research Council Canada \\
          \texttt{saif.mohammad@nrc-cnrc.gc.ca}}

\abstract{Over the last decade, Twitter has emerged as one of the most influential 
forums  for social, political, and health discourse. 
In this paper, we introduce a massive dataset of more than 45 million geo-located tweets
posted between 2015 and 2021
from US and Canada (\textit{TUSC}), especially curated for natural language analysis. 
We also introduce \textit{Tweet Emotion Dynamics (TED)} --- metrics to capture patterns of emotions associated with tweets over time.
We use TED and TUSC to explore 
the use of emotion-associated words 
across US and Canada; 
 across 2019 (pre-pandemic), 2020 (the year the pandemic hit), and 2021 (the second year of the pandemic); and across individual tweeters. 
 We show that Canadian tweets tend to have higher valence, lower arousal, and higher dominance than the US tweets.
 Further, we show that the COVID-19 pandemic had a marked impact on the emotional signature of tweets posted in 2020, when compared to the adjoining years. Finally, we 
 determine metrics of TED for 170,000 tweeters to benchmark characteristics of TED metrics at an aggregate level.
TUSC and the metrics for TED will enable a wide variety of research on studying how we use language to express ourselves, persuade, communicate, and influence, with particularly promising applications in public health, affective science, social science, and psychology.\\ 
\newline  }

\begin{document}

\maketitleabstract

\section{Introduction}

 \setitemize[0]{leftmargin=5mm}
 \setenumerate[0]{leftmargin=*}

\noindent  Over the last decade, Twitter has emerged not only as one of the most influential 
micro-blogging platforms, but also one of the most actively engaging (if sometimes polarizing) fronts 
  for social, political, and even health discourse.  Early work \cite{pak2010twitter,dodds2011temporal} identified tweets as a crucial indicator of 
public sentiment. Since then, 
various samples of tweet data have been used to analyze a wide variety of phenomena,
including the recent COVID-19 pandemic. 
However, past work largely uses
 topic-based keywords to obtain datasets of interest (often at the expense of geo-location information); for example, work that analyzes
 emotions in tweets that mention COVID-19-associated terms \cite{DBLP:journals/corr/abs-2004-03688,info:doi/10.2196/19447}. 
Further, very little work explores changes in patterns of emotions of individuals over time.

This paper introduces a new framework to analyze patterns of emotions associated with tweets over time, which we refer to as \textit{Tweet Emotion Dynamics (TED)}.
TED builds on ideas first introduced in \newcite{hipson2021emotion}, and applies metrics such as \textit{home base, variability,} and \textit{rise rate} to tweets.
We also introduce a new dataset of geo-located English \textit{\underline{T}weets from \underline{US} and \underline{C}anada} \textit{(TUSC)}.
 TUSC is not restricted to specific topics and so can be used to study tweets in general, as well as to study notable phenomena
(such as a pandemic, climate change, or polarizing political events) on tweets at large (as opposed to examining tweets directly discussing those phenomena).
TUSC also includes a subset, (\textit{TUSC100}), made up of tweets from 170K tweeters who each posted at least a 100 tweets between 2020 and 2021. TUSC100 is especially well suited for longitudinal analysis. 
The creation of the datasets included careful post-processing to 
make the resource particularly suitable for textual analysis. 

TUSC and TED can each be used, together or independently, to explore a wide range of research questions pertaining to tweets and
emotions that may be of interest to researchers in Psychology, Affective Science, Social Science, Behavioural Science, Public Health, NLP, and Linguistics.
In this paper, we use them to explore questions about how people use emotion-associated words in English tweets from US and Canada.\footnote{Words have associations with emotions; e.g., {\it success} with pleasure, {\it illness} with displeasure, etc.} 
We record the common characteristics of emotion word usage
from 2015 to 2021, with a special focus on 2020 --- the year that the WHO declared the Novel Coronavirus Disease (COVID-19) outbreak to be a pandemic --- 
and its adjoining years (2019 and 2021).
Finally, we benchmark individual tweeter behaviour in terms of various 
TED metrics.
Recording this information holds considerable promise in future work; 
for example, for studying the 
 emotional impact of the pandemic, for helping clinicians and patients track emotional well-being before and after health interventions, 
 studying emotion regulation and coping strategies, etc. 
The data (tweet IDs), Emotion Dynamics code, and visualizations are freely available through the project homepage.\footnote{\href{https://github.com/Priya22/EmotionDynamics}{https://github.com/Priya22/EmotionDynamics}} 

\section{Related Work}
\noindent We group related work into two kinds: 1. psychological and psychology-inspired research on the theory of emotions
and utterance emotion dynamics; and 
2. NLP research in analyzing emotions in tweets. 

\subsection{Emotions}

\noindent Several influential studies have shown that the three most fundamental, largely independent, dimensions of affect and connotative meaning are valence (V) (positiveness--negativeness / pleasure--displeasure), arousal (A) (active--sluggish), and dominance (D) (dominant-–submissive / in control--out of control) \cite{Osgood57,russell1977evidence,russell2003core}. Valence and arousal specifically are commonly studied in a number of psychological and neuro-cognitive explorations of emotion.

The NRC VAD Lexicon \cite{vad-acl2018} contains about twenty thousand commonly used English words (lemmas and common morphological variants) that have been scored on valence (0 = maximally unpleasant, 1 = maximally pleasant), arousal (0 = maximally calm/sluggish, 1 = maximally active/intense), and dominance (0 = maximally weak, 1 = maximally powerful).\footnote{\url{http://saifmohammad.com/WebPages/nrc-vad.html}} As an example, the word \textit{nice} has a valence of .93, an arousal of .44, and dominance of .65, whereas the word despair has a valence of .11, an arousal of .79, and dominance of .25.

\newcite{hipson2021emotion} introduced Utterance Emotion Dynamics (UED), a framework to quantify patterns of change of emotional states
associated with utterances
along a longitudinal (temporal) axis. Specifically, they proposed a series of metrics, including:\\[-16pt]
\begin{enumerate}
    \item \textit{Density or Mean}: A measure of the average utterance emotional state. For example, the proportion of emotion words a person utters over a given span of time. If each word has a real-valued emotion score (say, for V, A, and D), then this is calculated as the mean of emotion scores of the words in the utterance window. This roughly captures the utterance emotional state.\\[-18pt] 
    \item \textit{Variability}: The extent to which a speaker's utterance emotional state changes over time (measured as the standard deviation of the 
    emotion states). \\[-18pt]
    \item \textit{Home Base:} A speaker's home base is the subspace of high-probability emotional states where they are most likely to be found. This is formulated as the range of values within one standard deviation of the average of the emotion states at each timestep.\\[-18pt] 
    \item \textit{Rise and Recovery Rates:} Sometimes a speaker moves out of their home state, 
    reaches a peak value of emotion state, before returning to the home state. The rise rate quantifies the rate at which a speaker moves towards the peak; 
    recovery rate is the rate at which they go from the peak to the home state.\\[-16pt]
\end{enumerate}
One can determine UED metrics using:
1.\@ the utterances by a speaker, 
2.\@ the temporal information about the utterances, for e.g., time stamps associated with the utterances, or simply an ordering of utterances by time, 
and 3.\@ features of emotional state drawn from text. 
The emotional state at a particular instant 
can be determined using simple lexical features (say, drawn from emotion lexicons), predictions of supervised machine learning systems, etc.


\newcite{hipson2021emotion} apply this framework to utterances from a corpus of movie dialogues, which are naturally ordered along a temporal axis. 
They represented emotional state in a two-dimensional valence--arousal space.
The co-ordinates are determined by the average valence and arousal scores of the words (using the NRC VAD lexicon)
 in a small window of recent utterances (usually spanning 20 to 50 words). 
 Rolling windows of words (moving forward one-word at a time) determine the sequence of emotional states.
Here, we apply that framework to tweets. However, in this work we consider each of the valence, arousal, and dominance dimensions
separately (separate one-dimensional axes).
\subsection{Analyzing Emotions in Tweets}
\noindent \newcite{dodds2011temporal} analyzes large amounts of Twitter data 
to explore temporal patterns of `societal happiness'. 
\newcite{larsen2015we} 
show a correlation between patterns of emotional expression in tweets with WHO data on anxiety and suicide rates, across geographical location. \newcite{snefjella2018national} analyze differences in language use in 40 million tweets from Canada and the USA, and find that the former tend to use more positive language, which correlates with national character stereotypes of Canadians being more agreeable and less aggressive.
Twitter data has been used to study people's emotions during significant events, commonly revolving around certain tragedies and natural disasters, and significant political events. \newcite{doi:10.1177/0956797614562218} studied the changes in intensity of emotions of anxiety, anger, and sadness expressed on Twitter regarding the Sandy Hook Elementary School shooting. The 2016 US Presidential Election spurred several studies on the language used across geographical and political lines \cite{DVN/PDI7IN_2016}. Twitter was also used to measure the impacts of the COVID-19 pandemic on the emotional states and mental health of tweeters \cite{DBLP:journals/corr/abs-2004-03688}. \newcite{info:doi/10.2196/19447}, for example, looked at changes in the usage of tweets that expressed fear, anger, sadness, and joy in COVID-associated tweets from January 28 to April 9 2020.
In our work, we focus  on the emotion dimensions of valence, arousal, and dominance, rather than categorical dimensions such as anger, fear, sadness, etc. We also study these patterns of emotion usage across a large time period (2015--2021), and in geo-located tweets. 

\begin{table*}[t!]
{\small
    \centering
    \begin{tabular}{lrrr r rrr}
    \hline
            \textbf{Dataset}           & \multicolumn{3}{c}{\textbf{Canada}} &                                      & \multicolumn{3}{c}{\textbf{USA}}   \\
                            & \#tweets  & \# tweeters  &Av.TpT   & & \#tweets  & \# tweeters & Av.TpT \\
    \hline 
     {TUSC-Country}  &  &\\
 \rowcolor{gray!8}    $\;\;\;$ 2015 & 89,566 & 40,290 & 15.729 && 131,330 & 104,670 & 13.805\\
   $\;\;\;$ 2016  & 93,280 & 40,994 & 16.164 && 133,413 & 109,110 & 14.305\\
 \rowcolor{gray!8}      $\;\;\;$ 2017  & 94,364 & 39,258 & 18.067 && 133,854 & 107,080 & 16.015\\
   $\;\;\;$ 2018  & 95,403 & 38,866 & 21.763 && 133,066 & 105,227 & 19.394 \\
 \rowcolor{gray!8}       $\;\;\;$ 2019  & 330,361 & 70,122 & 22.040 && 339,186 & 204,311 & 19.341\\
 \rowcolor{yellow!8} $\;\;\;$ 2015--2019  & 702,974 & 159,284 & 18.753 && 870,849 & 516,885 & 16.572\\ \vspace*{-3mm}
 &\\
      $\;\;\;$ 2020  & 321,176 & 57,465 & 22.123 && 503,976 & 250,080 & 19.698\\
 \rowcolor{gray!8}   $\;\;\;$ 2021  & 304,106 & 49,128 & 22.192 && 478,798 & 214,653 & 19.566\\
  \rowcolor{yellow!8}    $\;\;\;$ 2015--2021  & 1,328,256 & 206,691 & 19.73 && 1,853,623 & 802,369 & 17.45 \\
  \vspace*{-3mm}
  &\\
{TUSC-City}     & & \\
       $\;\;\;$ 2020 (Apr--Dec)   & 15,039,503 & 716,063 & 19.275 && 23,470,855 & 2,669,081 & 17.556\\
  \rowcolor{gray!8} $\;\;\;$ 2021   & 22,371,990 & 798,602 & 19.367 && 43,693,643 & 3,247,124 & 17.306\\
\rowcolor{yellow!8}        $\;\;\;$ 2020--2021  & 37,411,493 & 1,049,774 & 19.327 && 67,164,498 & 4,274,374 & 17.413\\
     \hline 
    \end{tabular}
    \caption{\#tweets, \#tweeters, and Average number of token per tweet (Av.TpT) in the TUSC Datasets.}
    \label{tab:tusc-stats}
}
\end{table*}

\section{Tweets Dataset: TUSC}


\subsection{Sampling Tweets}
\label{sec:collection}

\noindent Twitter's regular API allows one to obtain a random sample of tweets from the past week.
However, the search is limited to only the tweets from the past week. 
The Academic search API provides access to historical tweets, but with a lower rate limit. 
To benefit from both APIs and to confirm that our results are 
consistent regardless of the API and search method,
we compiled two separate tweet datasets using each of the APIs:\\[3pt] 
\noindent 1. using Twitter's free API and its geo-location and random-sample switches to
collect tweets from 46 prominent American and Canadian cities. Data collection began in April 1, 2020 and is ongoing.
We refer to the dataset created with this method as \textit{TUSC-City}.\\[3pt] 
\noindent 2. using Twitter's Academic API to collect tweets 
emanating from US and Canada from Jan 2015 to Dec 2021.
The Academic API provides switches to specify the country of origin and the time span of search. However, the sample of results it provides tend to be in reverse chronological order for the specified time span.  
Thus, to obtain a sample of tweets from various time spans across the various years of interest, we employed the following strategy:
For each year of interest, we randomly generated a date time (using unix epoch seconds). We then specified a search interval of 8 hours starting from that date time. 
We repeated this procedure thousands of times for each year. Since we were especially interested in the years of 2019, 2020, and 2021, we collected more data from these years.
We refer to the resulting dataset as \textit{TUSC-Country}.

\subsection{Tweet Curation}
\noindent We curated the tweet collection to make it more suitable for computational natural language analyses by
applying the following steps:\\[-16pt]
\begin{itemize}
    \item Kept one tweet per user, per day. This mitigates the impact of highly prolific tweeters 
    and commercial accounts on the dataset.\\[-16pt]
    \item Kept only English language tweets (since the English set is the focus of this project). These are identified by the \texttt{iso\_language} tag provided by Twitter for each tweet.\\[-16pt] 
    \item Removed all retweets.\\[-16pt] 
    \item Removed all tweets containing a URL and/or links to media (to focus on textual tweets). 
    This also limits tweets by commercial organizations.\\[-16pt]
    \item Discarded all tweets with less than three tokens. This eliminates certain formulaic tweets such as wishes for holidays. The tweet text is tokenized using the Python implementation\footnote{\url{https://github.com/myleott/ark-twokenize-py}} of the Twokenizer package \cite{gimpel2010part,owoputi2013improved}.\\[-16pt]
\end{itemize}
\noindent We kept quotes and replies as they include new textual information.

\subsection{Key Data Statistics and Distribution}
\noindent We organize the TUSC tweets as per the sampling strategy used to obtain them (see TUSC-Country and TUSC-City in \S \ref{sec:collection})
as well as the year of posting (2015 through 2021), and country of origin (US, Canada). 
Table \ref{tab:tusc-stats} shows the number of tweets, number of tweeters, and average number of tokens per tweet in each of these dataset groupings.
(Table \ref{tab:stats-det} in the Appendix shows a breakdown by city for TUSC-City.)
It is interesting that an average Canadian tweet has about two more tokens per tweet than a US tweet (one possible explanation is the tendency of American tweeters to use more informal and non-standard language, as found in \newcite{snefjella2018national}).

TUSC-City is the larger dataset, and contains millions of tweets for many of the 46 cities for 9 months in 2020 (Apr--Dec), and all the months of 2021. 
It is useful for analyzing trends at the city--level, and also at the user-level, since we are more likely to have a large number of tweets from the same user. 

\section{Emotion Word Usage in American and Canadian Tweets}

\noindent The TUSC datasets can be used to answer several important questions about emotion word usage
in English tweets from US and Canada, including:
\vspace*{-2mm}
\begin{itemize}
    \item Are there notable trends across years in the valence, arousal, and dominance of tweets?
    Are we tweeting with more positive words, more negative words, more high arousal words, etc. than in past years?\\[-18pt]
    \item How has the COVID-19 pandemic impacted the emotionality of our tweets? 
    At what point of time in the pandemic did we use the most amount of words conveying 
    a lack of control and uncertainty? 
    How were individual cities impacted?\\[-18pt]
    \item How are Canada and US different in terms of emotion word usage? 
    Did the pandemic impact the emotionality 
    differently in the two countries?\\[-18pt]
\end{itemize}
\noindent We will explore these, and other, questions below. 

We used the NRC Valence, Arousal, and Dominance (NRC VAD) Lexicon \cite{vad-acl2018} to determine the emotion associations of the words  in  tweets.\footnote{\url{http://saifmohammad.com/WebPages/nrc-vad.html}} 
Specifically, we used the subset 
with entries for only the polar terms: i.e., 
only those valence entries 
that had scores $\leq$0.33 (negative words) or scores $\geq$0.67 (positive words).\footnote{There is no ``correct'' threshold to determine these classes; different thresholds simply make the positive and negative classes more or less  restrictive.}   
Similarly, only those arousal and dominance entries were included that had scores $\leq$0.33 or $\geq$0.67.
The entries with scores  between 0.33 and 0.67 are considered neutral for that dimension.\\[3pt]  
\noindent \textbf{Methodological Note 1:} As is good practise in lexicon-based analysis \cite{mohammad2020practical},
we removed lexicon entries for a small number of words  that were highly ambiguous (e.g., \textit{will, like})
or were expected to be frequently used in our tweets in a sense that is different from the usual predominant sense of the word (e.g., 
\textit{trump}).
The list of the 23 terms removed from the lexicon, and a description of the process of discovering them, is available in the Appendix.\\[3pt]
\noindent \textbf{Methodological Note 2:} Similar analyses can also be performed using categorical emotions, such as joy, sadness, fear, anger, etc., using the NRC Emotion Lexicon \cite{MohammadT10,Mohammad13}.\footnote{\url{http://saifmohammad.com/WebPages/NRC-Emotion-Lexicon.htm}}
See discussions on categorical and dimensional emotions in \cite{mohammad2020survey}.

\subsection{Average V, A, and D Across US--Canada}

 \begin{figure}[t!]
  \begin{center}
	\includegraphics[width=0.8\columnwidth]{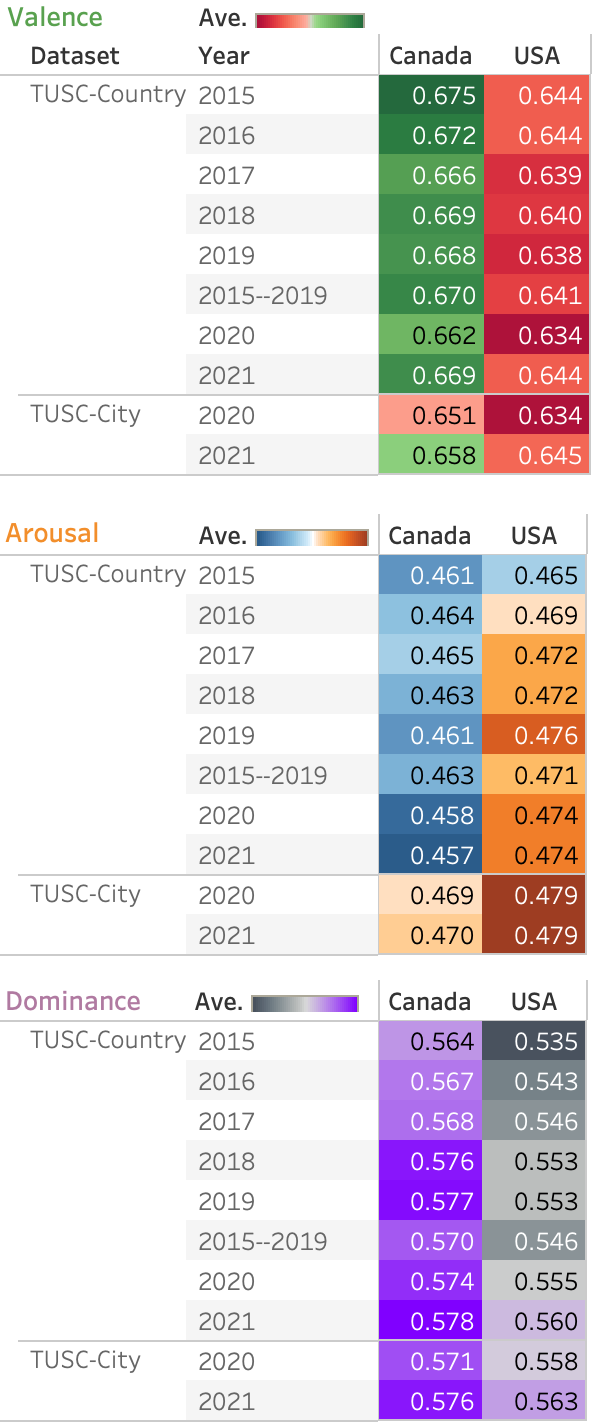} 
 	\caption{Average Valence, Arousal, and Dominance of words per tweet for each dataset (year).}
 	\label{fig:avg-vad}
 \end{center}
 \vspace*{-3mm}
 \end{figure}

\noindent For each tweet, we take the average of the  
valence, arousal, and dominance
values of each of the words in the tweet text. 
The averages are computed for TUSC-City over all tweets from each city, and for TUSC-Country at the country-level.  
We test whether the differences in values between countries and years are statistically significant by using the paired t-Test, with the significance threshold for the $p$-value set to 0.001.\\[4pt] 
\noindent \textbf{Yearly Trends:}
Figure \ref{fig:avg-vad} shows the average V, A, and D scores of tweets when aggregated at the country and year level, for 
the various data subsets. The gradient bars at the top show the colors used to indicate where the values lie in the spectrum from lowest to highest.\\[3pt] 
\noindent \textit{Valence:} 
Observe that the average valence of Canadian tweets is consistently higher (more positive) than the tweets from the US (statistically significant); the difference 
are steady across years. 
There is a slight downward trend for valence in both countries from 2015 to 2019.
We see the lowest values of mean valence occur in 2020 for both TUSC-Country and TUSC-City (the year the pandemic hit) for both the US and Canada. 
Average valence rises back up 
in 2021.\\[3pt] 
\noindent \textit{Arousal:} Overall, tweets from Canada have lower average arousal (more calm, less active) than the US (statistically significant).
Again, the difference in mean 
between the two countries remains relatively steady across years.
Across years, arousal values for both countries increase from 2015 till 2017; they then drop steadily for Canada, while the USA sees a peak in 2019 followed by slight drops in subsequent years.
  \begin{figure}[t]
	\includegraphics[width=\columnwidth]{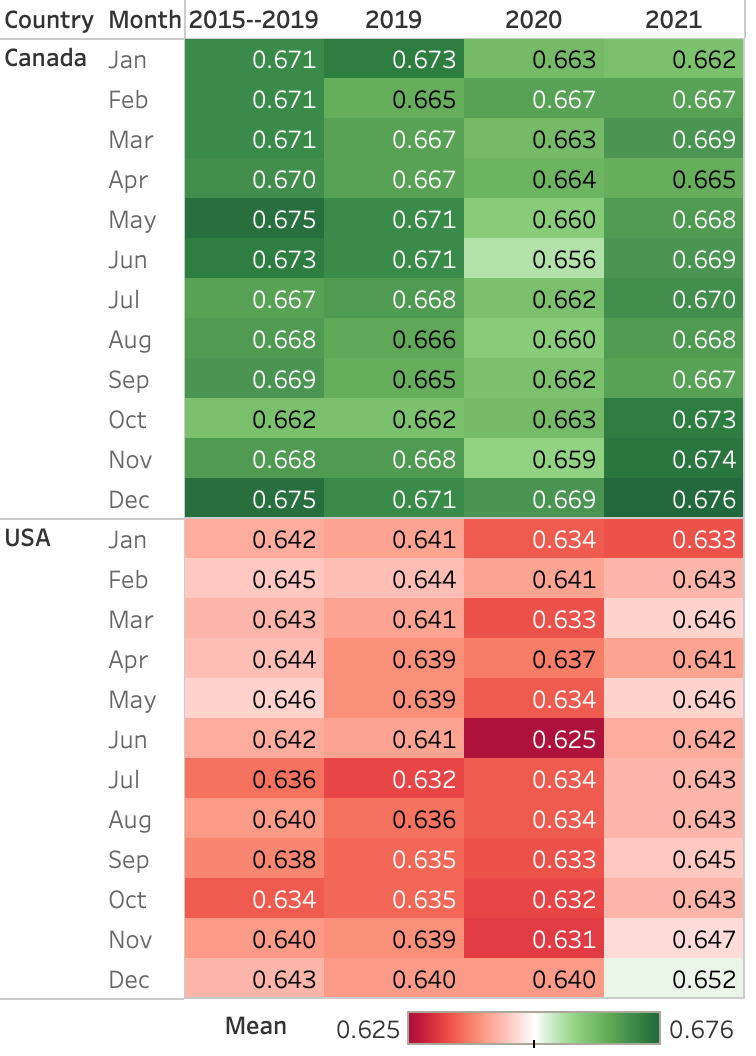}
 	\caption{Monthly trends in valence {\small (TUSC-Country)}.}
 	\label{fig:v-month}
\vspace*{-3mm}
 \end{figure}

\noindent \textit{Dominance:} 
Canada on the whole consistently has higher dominance values (greater feeling of control) than the US across the 
Both countries have the lowest dominance values in 2015, and the highest in 2021.
For all three dimensions, we note that the yearly trends observed in TUSC-Country are largely also observed in the TUSC-City trends, 
across 2020--2021.\\[6pt]
\noindent {\bf Monthly Trends:} Figure \ref{fig:v-month} shows a breakdown of the average valence scores 
at the month level, across years. (Figures \ref{fig:a-month} and \ref{fig:d-month} in the Appendix show the breakdown for arousal and dominance.)

We immediately notice from the color shading that Canada consistently has higher valence (green), lower arousal (blue), and higher dominance (purple) than the US, across the months and years.  
June 2020 is particularly notable as it has the lowest values of valence for both USA and Canada; we hypothesize that this is an effect of both the COVID-19 pandemic (the seriousness of which was starting to become evident a couple of months earlier in March 2020)
and the \textit{black lives matter} protests (which peaked after the Dylan Roof shooting incident). By contrast, the final months of 2021 have the highest positivity. This could be attributed to feelings of a potential return to normalcy, and 
a general uptick in mood due to the holiday season (this was just before the Omicron variant took root in early 2022). 

The dominance numbers indicate that April and May of 2020 for Canada and the USA are marked by some of the lowest scores, suggestive of a feeling of loss of control 
due to the onset of the global COVID-19 pandemic.

\subsection{Tweets with Emotional Terms}
\begin{figure}[t]
  \begin{center}
	\includegraphics[width=0.9\columnwidth]{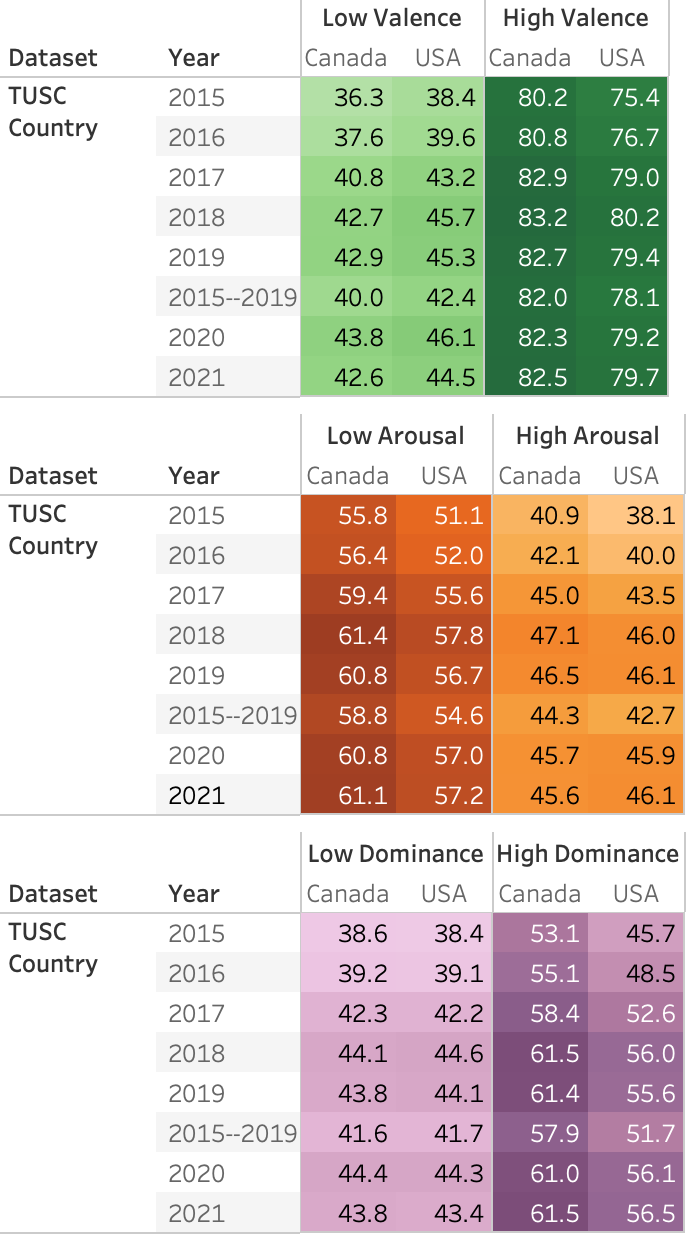} 
 	\caption{Percentage of tweets with at least one --- low valence word, high valence word, low arousal word, high arousal word, low dominance word, high dominance word --- across datasets (years).}
 	\label{fig:vad-perc}
 \end{center}
\vspace*{-6mm}
 \end{figure}

\noindent The experiments above showed notable differences in the average V/A/D scores of tweets across US and Canada. However, they also lead to further questions such as whether the higher valence in Canadian tweets is because of a greater usage of positive words or a lower usage of negative words. To explore such questions, 
we determine how frequently people post tweets with at least one high valence word,
how frequently people post tweets with at least one low valence word,
how frequently people post tweets with at least one high arousal word,
and so on.
High and low categorization of a word is based on whether their score (for V/A/D) is $\geq$0.67
or $\leq$0.33, respectively. Figure \ref{fig:vad-perc} shows the results. 

The darker shades of the color indicate a greater percentage of tweets had at least one of the relevant emotional words.\\[3pt]
Observe that in both American and Canadian tweets:\\[-16pt]
\begin{itemize}
    \item  people post markedly more tweets with at least one positive word than tweets with at least one negative word (about 100\% more).\\[-16pt]
    \item people post markedly more tweets with at least one low arousal word than tweets with at least one high-arousal word (about 40\% more).\\[-16pt]
    \item people post markedly more tweets with at least one high dominance word than tweets with at least one low-dominance word (about 33\% more). 
\end{itemize}

In terms of their differences, we see that the tweets from Canada are marked by both a higher usage of high-valence words, as well as a lower usage of low-valence words, than the US (statistically significant). 
Tweets from Canada have a higher proportion of low arousal words, whereas high arousal word usage is similar in both countries. 
Canadian tweeters use about the same number of low dominance words as those in the US, but use a greater number of high dominance words. 


Across years, low valence words increase in usage relatively steadily until 2020, and drop in 2021. For all dimensions, the sharpest rise in usage occurs from 2016 to 2017.
When comparing TUSC-Country 2020 with 2021, observe that the higher number of low valence words used is more prominent than the lower number of high valence words --- thus, the drop in average valence in 2020 (Figure 1) is because people \textit{tweeted more negative words}
(and not because people \textit{tweeted less positive words}).

See \newcite{VM2022-TED} for an analysis of the topics driving the differences across V, A, and D: across the years, and across US and Canada.

\section{Tweet Emotion Dynamics}


\noindent While the previous section looked at samples of tweets emanating from countries and cities as a whole,
in this section we explore individual tweeter behaviours and \textit{metrics} of emotion word usage over time.
Specifically, we apply the framework of Utterance Emotion Dynamics (UED) \cite{hipson2021emotion} to tweets to explore a number of questions, such as:\\[-16pt]
\begin{itemize}
    \item What is the usual range and distribution of various metrics of tweet emotion dynamics (TED), such as mean and recovery rate, for American and Canadian tweeters? Establishing benchmarks for these metrics is crucial for subsequent studies that may explore, for example,
    the impact of a health intervention on one's TED metrics. 
    \item Are there notable differences in the distributions of TED metrics across American and Canadian tweeters?\\[-16pt]
    \item Are there notable differences in the distributions of TED metrics across 2020 and 2021?\\[-12pt]
\end{itemize}

Recall that computing UED metrics requires: 1.\@ a set of texts associated with each speaker, 2.\@ a temporal ordering of these texts,
and 3.\@ a way to determine emotions associated with the texts.
The timestamp of a tweet provides the temporal order. The tweets from each speaker are then concatenated together and tokenized to obtain a ordered list of tokens. 
Next, a rolling window of 20 tokens is considered to determine the average V, A, and D scores of the words in that window. These scores are a representation of utterance emotional state corresponding to that window.
The rolling window is moved forward one word at a time to determine the subsequent averages.\footnote{Variations of this approach that do not use rolling windows across tweet boundaries produce similar results.}  
In the rest of this section, we use the term \textit{mean} to refer to the mean of all the rolling window averages for a speaker.



We re-implemented the \newcite{hipson2021emotion} R code in Python. Also, rather than using the valence--arousal two-dimensional (ellipse-based) representation of emotions explored in their work, we analyse the dynamics of each emotion dimension separately along one-dimensional axes. This eases the interpretation of the UED metrics, and allows for considering more dimensions such as dominance.
We also break down the rise and recovery rates into separate rates corresponding to when one is moving from the home base
to higher emotion values (Hm--Hi), from the highest value to home (Hi--Hm), from the home base
to lower emotion values (Hm--Lo), and from the lowest value to home (Lo--Hm).
This allows us to explore whether, for example, some tweeters have close-to-median Hi--Hm rates, but markedly low Lo--Hm rates (indicating that once they start uttering negative words, then they tend to dwell in the negatives and only gradually return to their home state). 

   \begin{figure}[t]
	\includegraphics[width=\columnwidth]{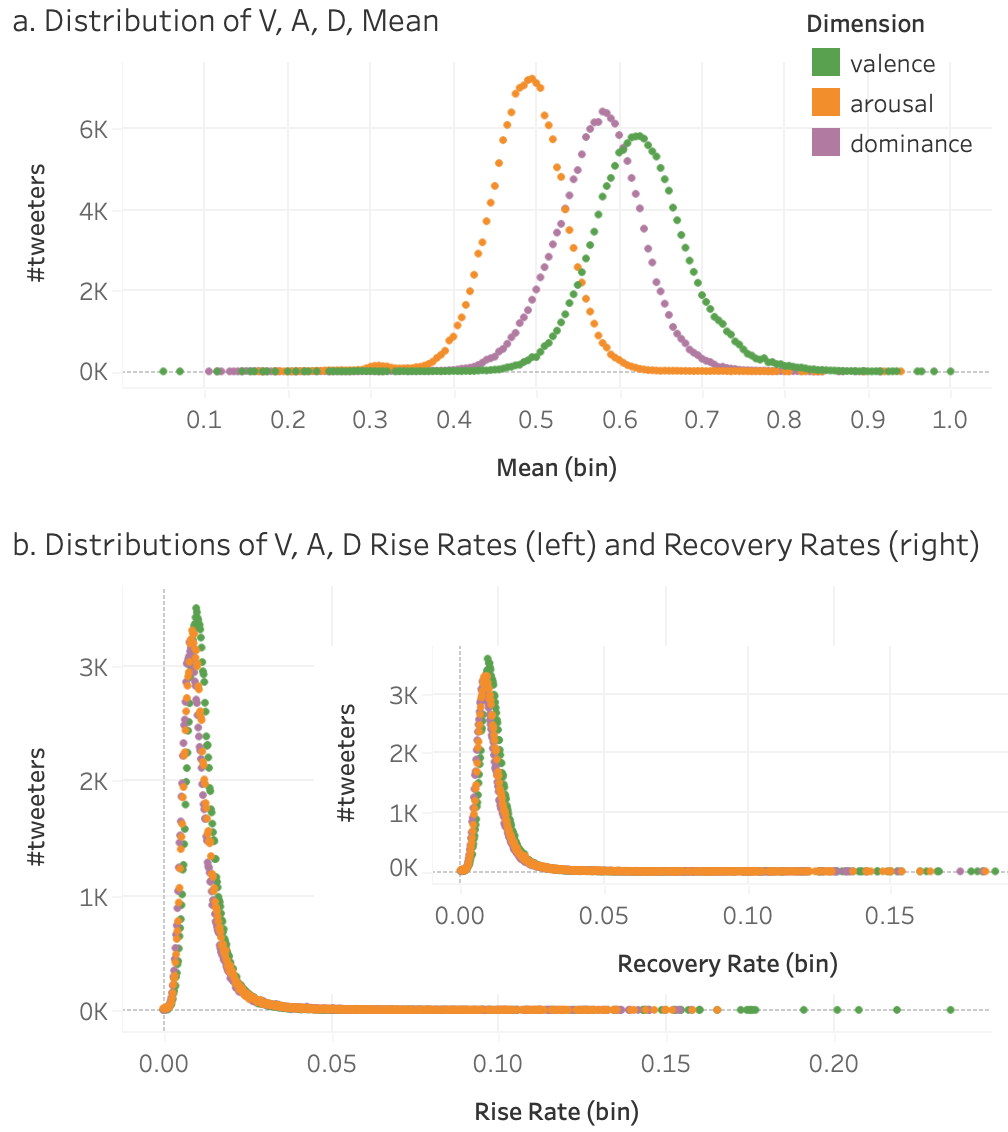}
 	\caption{Distributions of Means, Rise Rates, and Recovery Rates for Valence, Arousal, and Dominance (TUSC100).}
 	\label{fig:ted-hist}
 \vspace*{-3mm}
 \end{figure}
 
   \begin{figure*}[t]
  \begin{center}
	\includegraphics[width=1.7\columnwidth]{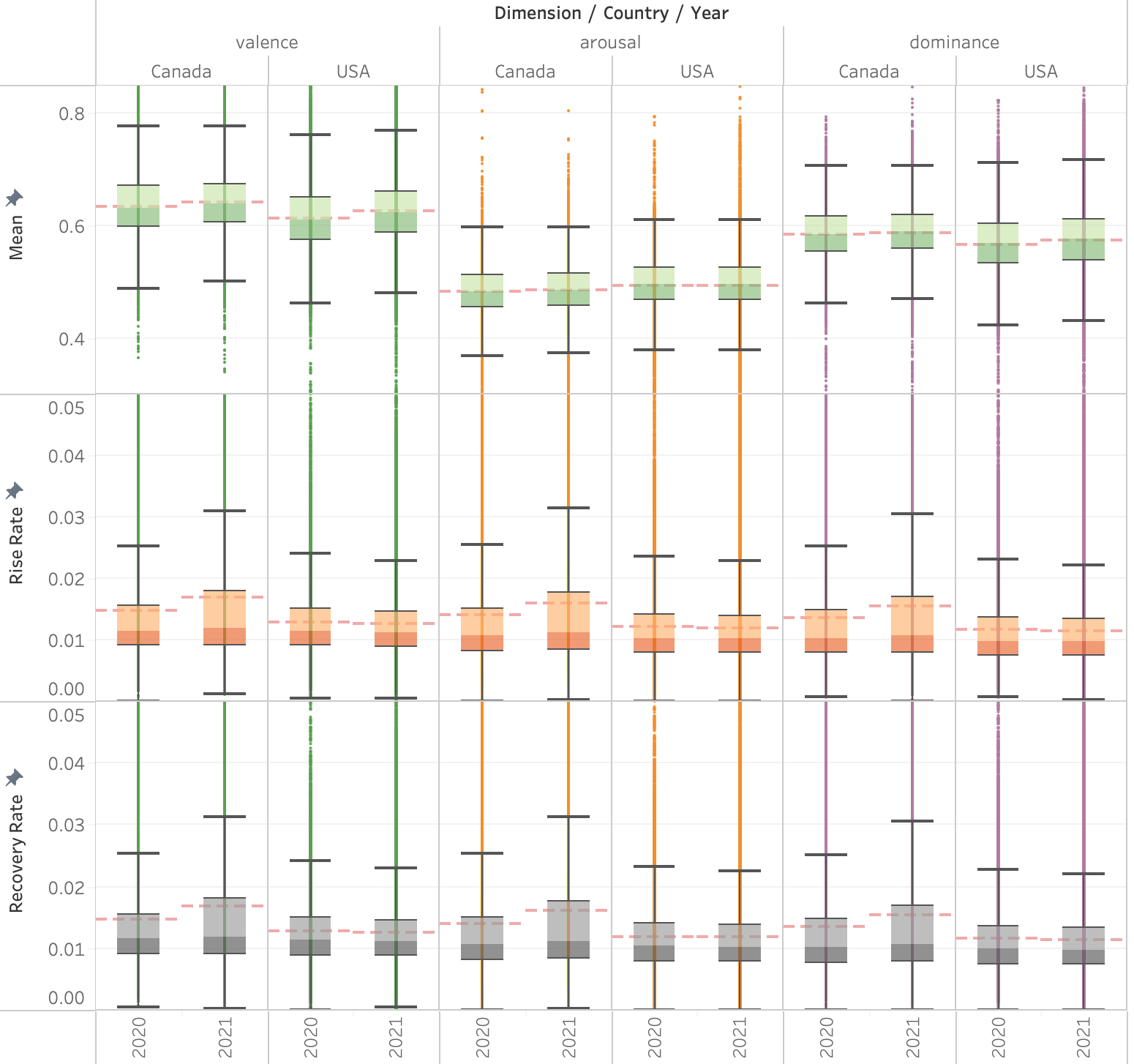}
 	\caption{Box plots of  means, rise rates, and recovery rates of Valence, Arousal, and Dominance of tweeters in 2020 and 2021 (TUSC100-2020 and TUSC100-2021).}
 	\label{fig:ted-box}
 \end{center}
\vspace*{-3mm}
 \end{figure*}
 

We use the code to determine TED metrics for the tweeters in  in the TUSC-Country dataset.  
Only tweeters with at least 100 tweets in a year were considered, since drawing inferences about one's tweeting behavior requires a sufficient sample size. 
There were about 40K such tweeters in the 2020 subset and about 130K such tweeters in the 2021 subset.
We refer to their tweets (5.6M from 2020 and 19M from 2021) as the \textit{TUSC100-2020} and \textit{TUSC100-2021} datasets, respectively. Average number of tweets by a tweeter in these datasets is 153 (no tweeters had more than 365 tweets due to our earlier stated `one tweet per user per day' pre-processing policy). See Table \ref{tab:t100-stats} in the Appendix for detailed statistics.

Figure \ref{fig:ted-hist} plots the distributions of some of the metrics for the joint set of 2020 and 2021 tweeters. 
The plots in (a) are 
distributions 
of the mean values for the three emotion dimensions (V, A, D). 
The x-axis is made up of bins of size 0.005 (from 0--0.005 to 0.995--1). 
The y-axis indicates the number of tweeters with mean values in each of the bins. 
Observe that the means for V, A, and D all follow a near-normal distribution. Mean valence  and dominance values are more spread out compared to arousal values.
For V, most fall between 0.5--0.8 , with a median value of around 0.65, though we can see that there is a long tail of outliers. Dominance scores are spread around a median score of 0.6, and the median is even lower for arousal (0.49).

Figure \ref{fig:ted-hist}(b) shows 
distributions for rises rates and recovery rates. 
Observe that these have a much narrower spread, and the distributions for all three dimensions are roughly the same.

Figure \ref{fig:ted-box} shows box and whisker plots of the same three metrics: mean, rise rate, and recovery rates.
However, separate plots are shown for tweeters from US and Canada, and across 2020 and 2021.
The shaded region (the box) indicates the ``middle portion" of the data distribution, i.e, the range covered between the first quartile (the 25\% mark) and the third quartile (the 75\% mark), with the median (50\% mark) lying at the border of the light and dark shaded regions. The whiskers, the lines on either end of the plot, are at a distance of 1.5 times the inter-quartile length (inter-quartile length is the distance between the first and third quartiles). Points beyond the whiskers are considered outliers.  Additionally, the average value (mean) is indicated with the pink horizontal dashed line.

Observe that the mean valence is lower in 2020 than in 2021, and Canadian tweeters on average use more positive words than their US counterparts.\footnote{These trends align with the trends observed in Table 2.}
The distributions for mean arousal are quite similar across  2020 and 2021, but US tweeters have slightly higher mean arousal values.  Canadian tweeters have a slightly higher median of dominance scores than US tweeters; whereas the US tweeters tend to have a wider range of dominance values.
The difference in the distributions of the mean values for Canada and US is statistically significant for all three dimensions ($p$-values $< 0.001$).

The median rise rates and recovery rates do not differ markedly across countries or years. 
However, there is a notably large range of the third quartile (the quartile above the median) 
for Canadian tweeters in 2021.
These are tweeters who are quicker to jump in and out of their home base. Tables \ref{tab:ted-val}, \ref{tab:ted-ar}, \ref{tab:ted-dom} of the Appendix report mean scores for all of the TED metrics, averaged across all tweeters by country and year.
This includes a breakdown of the rates into Hm-Hi, Hi-Hm, Hm-Lo, and Lo-Hm.
Notable trends there are that the average rise and recovery rates 
on the high side of
the home state (Hm-Hi, Hi-Hm) are lower than for the 
low side of the home state
(Lo-Hm, Hm-Lo), for the valence and dominance dimensions. This says that tweeters are slower to rise to more positive and more dominant states, but quicker to both descend to more negative and less dominant states, and recover from them; similarly, they are slower to transit to and from states of high activity (high dominance). This difference between Hi and Lo rates is reversed for arousal. Thus, we are quicker to rise to states of high arousal, and come back down from them to the home state. 

We also noticed in our analyses that there exist several tweeters 
that have very high rise rates but normative recovery rates, 
and also tweeters
that have very high rise rates but normative recovery rates.
Identifying such characteristics and tracking them in the context of health interventions is particularly promising future work.
However, it should be noted that we strongly encourage such studies, when conducted, to be led by clinicians and psychologists, with appropriate consent and ethics approvals. 

\subsection{City as Speaker}
\noindent An interesting variation of the experiments above, is to consider each city as a `speaker', rather than individual tweeters. 
Figure \ref{fig:city-speaker-mean}, in the Appendix, shows the average TED metrics for each of the 46 cities in TUSC-City. The color gradients make it easy to spot which cities have had markedly high V/A/D means across 2020 and 2021. 
Consistent with some of the earlier country-level results, we see that the Canadian cities tend to have higher valence, lower arousal, and higher dominance, than the US cities.
London, Ottawa, Halifax, and Victoria have the highest valence (most positive).
From the set of Canadian cities, Windsor stands out as an anomaly with valence close to many US cities. Detroit, Houston, Los Angeles, and Philadelphia have some of the lowest valence values of all cities. All cities improve from 2020 to 2021, some more drastically than others (Boston, Indianapolis, San Jose, for example). Quebec City and Windsor have the highest arousal rates in Canada; in the US, El Paso is at the top for both years. Nashville, San Francisco, San Jose, and Seattle have lower arousal rates (more in line with the average Canadian city).
Washington, San Jose, and Boston also show markedly high dominance, as well as San Francisco. Among Canadian cities, Ottawa and Victoria have the highest dominance scores for 2020 and 2021, and Windsor again the lowest.

Figure \ref{fig:city-speaker-rest}, in the Appendix, shows values of the variabilities, rise rates, and recovery rates for the valence dimension. Looking at the column for variability, Windsor jumps out among the Canadian cities for having comparatively higher variability. Washington and Phoenix in 2020 have relatively high variability. Moving to the next columns, Windsor again has the highest rise and recovery rates among Canadian cities; US cities are the on the whole quicker to rise and fall. 

The various metrics listed for various cities should be useful to those interested in the tweets from particular cities.
Future work will drill down further into the data for individual cities to determine the factors driving the emotion word usage.

\section{Conclusion}
\noindent We introduced the Tweet Emotion Dynamics (TED) framework to quantify changes in emotions associated with tweets over time. 
We also released TUSC  --- a large collection of English geo-located tweets from Canada and the USA that were posted between 2015 and 2021. 
We studied emotion word usage in this data, using multiple metrics, for the primary dimensions of valence, arousal, and dominance. Our results showed interesting trends in the emotions expressed by tweeters from the two countries across different years, and also uncovered contrasts between Canadian and US tweeters. 
An expanded version of this paper presents further experiments, including a deeper analysis of the words and topics driving the TED metrics
\cite{VM2022-TED}. 
Future work will explore tweets from other countries and also tweets in languages other than English. 
We will also examine how the TED framework can be useful to clinicians and psychologists for measuring mental health outcomes, at an aggregate level, from social media data. Additionally, we will study the applications of Utterance Emotion Dynamics in other contexts such as novels, personal diaries, forum posts, and speech.

\section*{Acknowledgements}
We thank Will Hipson for early discussions on his R code for utterance emotion dynamics. The first author is supported by funding from the Natural Sciences and Engineering Research Council of Canada, and resources provided by the Vector Institute of Artificial Intelligence. 

 \section{Bibliographical References}
\label{main:ref}

\bibliographystyle{lrec2022-bib}
\bibliography{references}

 \section*{Appendix}

\section{Ethics Considerations}
\noindent Emotions are complex, private, and central to our experience. There is also tremendous variability in how we
express emotions through language. Thus several ethical considerations are relevant to textual analysis of emotions.
Some that we would particularly like to highlight are listed below:
\vspace*{-2mm}
\begin{itemize}
    \item We only release the tweet IDs for each tweet, which will need to be hydrated by users of our dataset with the Twitter API. If any of the tweets are deleted by the associated tweeter, it will no longer be accessible.
\vspace*{-2mm}
    \item Our work on studying emotion word usage should not be construed as detecting how people feel; rather, we draw inferences on the emotions that are conveyed by users via the language that they use. 
\vspace*{-2mm}
    \item The language used in an utterance may convey information about the emotional state (or perceived emotional state) of the speaker, listener, or someone mentioned in the utterance. However, it is not sufficient for accurately determining any of their momentary emotional states. Deciphering true momentary emotional state of an individual requires extra-linguistic context and world knowledge.
    Even then, one can be easily mistaken.
\vspace*{-2mm}
    \item The inferences we draw in this paper are based on aggregate trends across large populations. We do not draw conclusions about specific individuals or momentary emotional states.
\vspace*{-2mm}
    \item We do not recommend the use of TED metrics to draw inferences about individuals, unless: 1. it is exercised with extreme caution, 
    2. for the express benefit, and with consent, of the people whose data is used, 3. the work is led by sbject-matter experts such as psychologists or clinicians, and 4. automatically drawn information is used as one source of information among many by human experts.
\vspace*{-2mm}
    \item Any information drawn from these metrics regarding one's language use should not be used to negatively impact the individual.
\end{itemize}
\vspace*{-2mm}
\noindent See \newcite{Mohammad22AER} for a detailed discussion on the ethical considerations of automatic emotion recognition and 
\newcite{mohammad2020practical} for practical and ethical considerations in the effective use of emotion lexicons.

\section{Additional Emotion Word Usage Statistics from TUSC}
\noindent We present, in this section, additional tables and figures that record details of emotion word usage broken down by city (for the 46 cities considered) and by month of year. Table \ref{tab:t100-stats} shows the number of tweets and tweeters in each subset of the TUSC100 dataset. Tables \ref{tab:ted-val}, \ref{tab:ted-ar}, and \ref{tab:ted-dom} tabulate the numbers corresponding to the plots in Figure \ref{fig:ted-box} in the main paper.

\section{Modified NRC VAD Lexicon}
\noindent When applying lexicon-based analyses to datasets from a specific domain, Mohammad \shortcite{mohammad2020practical} recommends updating the emotion lexicons to remove terms that can be used in a sense different from the predominant word sense. Since manual examination of all the words in a large dataset is difficult, this step is recommended for at least the frequent terms. 

For our analyses, we first compiled a list of all the terms from the NRC VAD lexicon that occurred in at least 0.1\% of the tweets from either Canada or the USA, for any of the years in the TUSC-Country dataset  (2015--2021). Both of the authors of this work 
examined the list and identified words that were highly ambiguous or occurred in the tweets predominantly in a sense different from 
what would be expected if people were shown the word out of context 
(as was the case of the original annotations in the NRC VAD lexicon). 
In all, 23 such words were identified (shown in Table \ref{tab:removed-terms}). 
The entries for these words were removed from the lexicon before conducting the experiments described in the paper.

To examine the impact of the above lexicon update, 
we repeated all the experiments with the unmodified lexicon as well. 
We observed that while the numerical values of the UED metrics changed slightly (as would be expected), all relative trends remained the same, across countries and across years.   
The interested reader can find the scores for the UED metric and the complete set of experiments with the unmodified lexicon in version 2 (an older version) of the paper on ArXiv.\footnote{https://arxiv.org/abs/2204.04862}

\begin{table}[t]
    \centering
    \begin{tabular}{llll}
    \hline
    have & will & one & high \\
    may & way & kind & be\\
    thing & things & number & seem\\
    do & look & three & third\\
    five & senate & say & talk\\ 
    president & trump & like & \\
    \hline

    \end{tabular}
    \caption{List of terms that were removed from the NRC VAD Lexicon.} 
    \label{tab:removed-terms}
\end{table}

\begin{table*}[ht!]
{\small
\begin{center}
\begin{tabular}{lrrrrrr}
\hline
                    &\multicolumn{2}{c}{\textbf{2020}}    &\multicolumn{2}{c}{\textbf{2021}}  \\
\textbf{City}       & \textbf{\# tweets} & \textbf{\# tweeters}  & \textbf{\# tweets} & \textbf{\# tweeters}\\
\hline

\textit{Canada}\\
$\;\;\;\;\;\;$ Brampton & 1,436,865 & 159,974 & 2,430,329 & 188,216\\
 $\;\;\;\;\;\;$ Calgary & 294,911 & 31,988 & 503,173 & 39,416\\
 $\;\;\;\;\;\;$ Edmonton & 806,116 & 43,427 & 1,319,950 & 49,058\\
 $\;\;\;\;\;\;$ Etobicoke & 1,318,119 & 157,429 & 2,379,928 & 191,653\\
 $\;\;\;\;\;\;$ Halifax & 572,562 & 23,733 & 678,033 & 23,541\\
 $\;\;\;\;\;\;$ Hamilton & 446,038 & 37,023 & 702,761 & 43,537\\
 $\;\;\;\;\;\;$ Laval & 453,670 & 48,145 & 733,844 & 58,344\\
 $\;\;\;\;\;\;$ London & 298,615 & 16,977 & 428,929 & 18,928\\
 $\;\;\;\;\;\;$ Mississauga & 450,835 & 97,328 & 977,517 & 142,817\\
 $\;\;\;\;\;\;$ Montreal & 627,159 & 52,396 & 1,048,093 & 64,363\\
 $\;\;\;\;\;\;$ North York & 1,274,462 & 148,271 & 1,685,201 & 152,105\\
 $\;\;\;\;\;\;$ Okanagan & 30,771 & 1,814 & 37,424 & 1,813\\
 $\;\;\;\;\;\;$ Ottawa & 1,055,035 & 55,430 & 1,332,680 & 56,621\\
 $\;\;\;\;\;\;$ Quebec & 284,665 & 16,380 & 377,342 & 18,100\\
 $\;\;\;\;\;\;$ Scarborough & 720,165 & 108,498 & 710,181 & 86,328\\
 $\;\;\;\;\;\;$ Surrey & 1,115,467 & 84,001 & 1,679,642 & 94,177\\
 $\;\;\;\;\;\;$ Toronto & 2,058,494 & 182,730 & 2,557,606 & 182,792\\
 $\;\;\;\;\;\;$ Vancouver & 402,418 & 53,655 & 634,307 & 63,561\\
 $\;\;\;\;\;\;$ Victoria & 340,720 & 14,787 & 436,905 & 15,565\\
 $\;\;\;\;\;\;$ Windsor & 443,712 & 58,545 & 893,922 & 72,975\\
 $\;\;\;\;\;\;$ Winnipeg & 608,704 & 27,954 & 824,223 & 29,365\\

\textit{US}\\
$\;\;\;\;\;\;$ Austin & 1,244,776 & 102,841 & 2,242,561 & 125,526\\
 $\;\;\;\;\;\;$ Boston & 764,257 & 100,276 & 1,641,142 & 130,145\\
 $\;\;\;\;\;\;$ Charlotte & 997,197 & 76,528 & 1,566,062 & 86,892\\
 $\;\;\;\;\;\;$ Chicago & 721,075 & 142,591 & 1,652,701 & 194,565\\
 $\;\;\;\;\;\;$ Columbus & 809,160 & 69,931 & 1,445,275 & 81,135\\
 $\;\;\;\;\;\;$ Dallas & 674,613 & 129,304 & 1,671,887 & 181,895\\
 $\;\;\;\;\;\;$ Denver & 1,198,813 & 98,697 & 1,712,785 & 106,959\\
 $\;\;\;\;\;\;$ Detroit & 749,506 & 77,560 & 1,418,484 & 94,202\\
 $\;\;\;\;\;\;$ El Paso & 692,705 & 38,096 & 781,937 & 37,335\\
 $\;\;\;\;\;\;$ Fort Worth & 1,649,842 & 188,443 & 2,794,053 & 215,169\\
 $\;\;\;\;\;\;$ Houston & 1,557,488 & 195,548 & 2,358,286 & 209,520\\
 $\;\;\;\;\;\;$ Indianapolis & 710,808 & 65,665 & 1,287,399 & 78,214\\
 $\;\;\;\;\;\;$ Jacksonville & 723,513 & 48,230 & 1,000,620 & 53,257\\
 $\;\;\;\;\;\;$ Los Angeles & 1,028,102 & 246,491 & 2,470,750 & 337,004\\
 $\;\;\;\;\;\;$ Memphis & 876,988 & 49,835 & 1,263,728 & 54,254\\
 $\;\;\;\;\;\;$ Nashville & 584,997 & 68,306 & 963,947 & 79,006\\
 $\;\;\;\;\;\;$ New York & 1,079,557 & 281,109 & 2,288,500 & 361,670\\
 $\;\;\;\;\;\;$ Philadelphia & 1,142,818 & 127,257 & 2,579,605 & 161,491\\
 $\;\;\;\;\;\;$ Phoenix & 531,390 & 81,093 & 1,454,042 & 117,145\\
 $\;\;\;\;\;\;$ San Antonio & 1,213,537 & 91,427 & 1,835,376 & 98,868\\
 $\;\;\;\;\;\;$ San Diego & 1,080,361 & 101,554 & 1,940,120 & 122,278\\
 $\;\;\;\;\;\;$ San Francisco & 1,123,356 & 132,023 & 2,179,371 & 159,727\\
 $\;\;\;\;\;\;$ San Jose & 790,511 & 72,049 & 1,184,999 & 78,750\\
 $\;\;\;\;\;\;$ Seattle & 940,092 & 114,848 & 1,968,319 & 141,016\\
 $\;\;\;\;\;\;$ Washington & 585,393 & 123,576 & 1,991,694 & 189,873\\
\hline
\textbf{All} & \textbf{38,510,358} & \textbf{3,332,189} & \textbf{66,065,633} & \textbf{3,976,481}\\
      \hline
\end{tabular}
\caption{The number of tweets and tweeters in TUSC-City for 2020 and 2021.} 
\label{tab:stats-det}
 \end{center}
 }
\end{table*}

\clearpage
\newpage


\begin{figure}[ht]
	\includegraphics[width=\columnwidth]{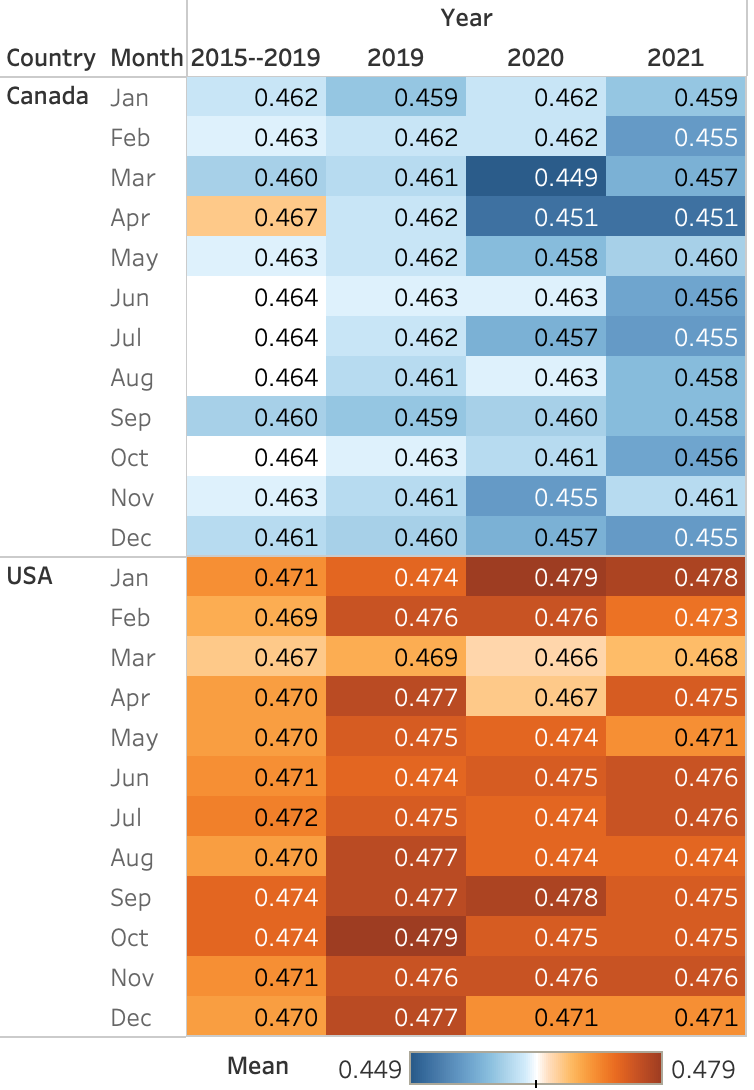}
 	\caption{Monthly trends in \textbf{Arousal} of tweets across years (TUSC-Country).}
 	\label{fig:a-month}
 \end{figure}
 
 \begin{figure}[t]
	\includegraphics[width=\columnwidth]{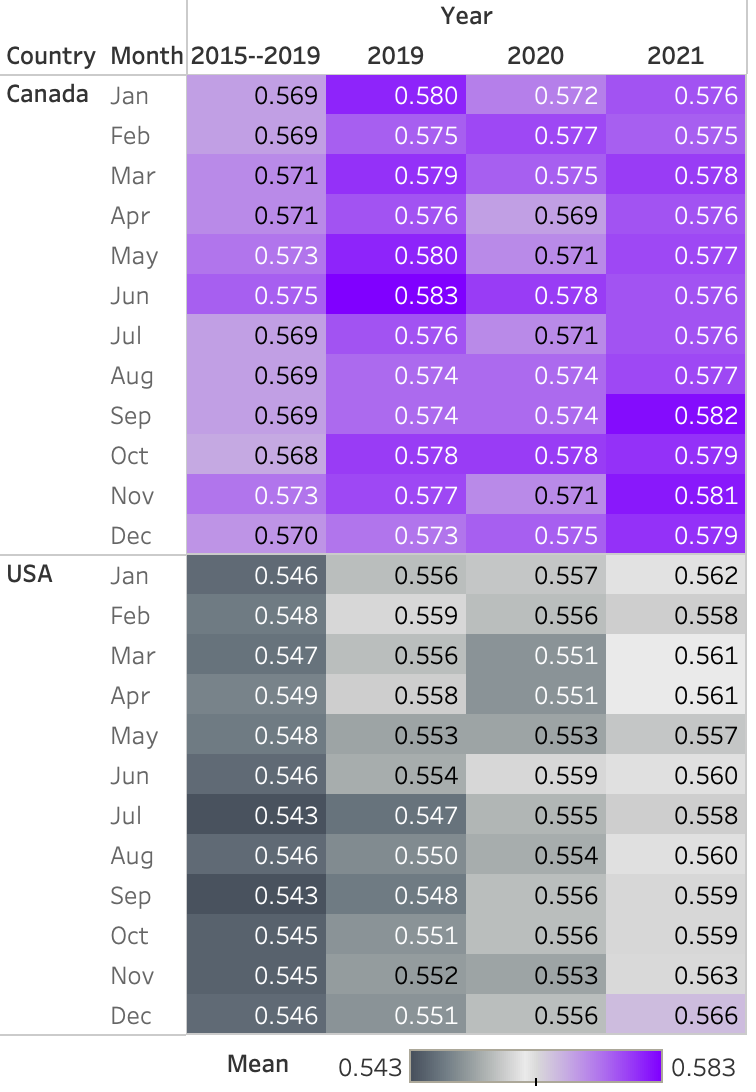}
 	\caption{Monthly trends in \textbf{Dominance} of tweets across years (TUSC-Country).}
 	\label{fig:d-month}
 \end{figure}
 


\clearpage
\newpage

\begin{table}[t]
    \centering
     {\small
    \begin{tabular}{lrr}
    \hline
    \textbf{Dataset} & \textbf{\# tweets} & \textbf{\# tweeters} \\ \hline 
      \textbf{2020}&\\[1pt]
$\;\;\;$ $\;\;\;$ Canada  &  3,038,530 & 20,887\\
$\;\;\;$ $\;\;\;$ USA  &  2,641,694 & 19,709\\
\textbf{2021}&\\[1pt]
$\;\;\;$ $\;\;\;$ Canada  &  7,467,446 & 45,573\\
$\;\;\;$ $\;\;\;$ USA  &  11,675,372 & 76,223\\
    \hline 
    \end{tabular}
    \caption{Number of tweets and tweeters in the TUSC100 dataset.}
    \label{tab:t100-stats}
    }
    \vspace*{35mm}
\end{table}

\begin{table}[t]
    \centering
  {\small
    \begin{tabular}{llrr}
    \hline
    \textbf{Data}  & \textbf{Year} & \textbf{Canada} & \textbf{USA}   \\\hline 
\textit{Mean} & 2020 & 0.6320 & 0.6132\\
  & 2021 & 0.6387 & 0.6257\\
[3pt]
\textit{Variability} & 2020 & 0.0708 & 0.0714\\
  & 2021 & 0.0700 & 0.0705\\
[3pt]
\textit{Rise Rate} & 2020 & 0.0121 & 0.0128\\
  & 2021 & 0.0117 & 0.0123\\
[3pt]
\textit{Recovery Rate} & 2020 & 0.0120 & 0.0127\\
  & 2021 & 0.0118 & 0.0123\\
[3pt]
\textit{Hm-Hi Rate} & 2020 & 0.0118 & 0.0129\\
  & 2021 & 0.0113 & 0.0121\\
[3pt]
\textit{Hi-Hm Rate} & 2020 & 0.0118 & 0.0129\\
  & 2021 & 0.0115 & 0.0122\\
[3pt]
\textit{Hm-Lo Rate} & 2020 & 0.0143 & 0.0149\\
  & 2021 & 0.0140 & 0.0145\\
[3pt]
\textit{Lo-Hm Rate} & 2020 & 0.0141 & 0.0148\\
  & 2021 & 0.0139 & 0.0144\\
[3pt]
     \hline 
    \end{tabular}
    \caption{Tweet \textbf{Valence} dynamics metrics of tweeters in TUSC100. Averaged across all tweeters (not considering the cities they came from).}
    \label{tab:ted-val}
    }
\end{table}


\vfill\null 
\columnbreak

\begin{table}[t!]
    \centering
     {\small
    \begin{tabular}{llrr}
    \hline
    \textbf{Data} & \textbf{Year}  & \textbf{Canada} & \textbf{USA}   \\
    \hline 







\textit{Mean} & 2020 & 0.4828 & 0.4935\\
  & 2021 & 0.4854 & 0.4932\\
[3pt]
\textit{Variability} & 2020 & 0.0599 & 0.0593\\
  & 2021 & 0.0599 & 0.0595\\
[3pt]
\textit{Rise Rate} & 2020 & 0.0116 & 0.0120\\
  & 2021 & 0.0113 & 0.0117\\
[3pt]
\textit{Recovery Rate} & 2020 & 0.0115 & 0.0119\\
  & 2021 & 0.0113 & 0.0116\\
[3pt]
\textit{Hm-Hi Rate} & 2020 & 0.0129 & 0.0130\\
  & 2021 & 0.0125 & 0.0129\\
[3pt]
\textit{Hi-Hm Rate} & 2020 & 0.0127 & 0.0130\\
  & 2021 & 0.0126 & 0.0128\\
[3pt]
\textit{Hm-Lo Rate} & 2020 & 0.0121 & 0.0127\\
  & 2021 & 0.0118 & 0.0123\\
[3pt]
\textit{Lo-Hm Rate} & 2020 & 0.0120 & 0.0125\\
  & 2021 & 0.0117 & 0.0121\\
     \hline 
    \end{tabular}
\caption{\textbf{Arousal} dynamics metrics of tweeters in TUSC100. Averaged across all tweeters (not considering the cities they came from).}
    \label{tab:ted-ar}
    }
\end{table}

\begin{table}[t]
    \centering
     {\small
    \begin{tabular}{llrr}
    \hline
    \textbf{Data} & \textbf{Year}  & \textbf{Canada} & \textbf{USA}   \\
    \hline 
\textit{Mean} & 2020 & 0.5821 & 0.5671\\
  & 2021 & 0.5873 & 0.5733\\
[3pt]
\textit{Variability} & 2020 & 0.0573 & 0.0556\\
  & 2021 & 0.0569 & 0.0557\\
[3pt]
\textit{Rise Rate} & 2020 & 0.0113 & 0.0116\\
  & 2021 & 0.0109 & 0.0113\\
[3pt]
\textit{Recovery Rate} & 2020 & 0.0112 & 0.0117\\
  & 2021 & 0.0109 & 0.0112\\
[3pt]
\textit{Hm-Hi Rate} & 2020 & 0.0114 & 0.0118\\
  & 2021 & 0.0111 & 0.0115\\
[3pt]
\textit{Hi-Hm Rate} & 2020 & 0.0114 & 0.0119\\
  & 2021 & 0.0111 & 0.0114\\
[3pt]
\textit{Hm-Lo Rate} & 2020 & 0.0127 & 0.0129\\
  & 2021 & 0.0124 & 0.0126\\
[3pt]
\textit{Lo-Hm Rate} & 2020 & 0.0126 & 0.0128\\
  & 2021 & 0.0123 & 0.0125\\
[3pt]

     \hline 
    \end{tabular}
    \caption{Tweet \textbf{Dominance} Dynamics metrics of tweeters in TUSC100. Averaged across all tweeters (not considering the cities they came from).}
    \label{tab:ted-dom}
    }
\end{table}

\vfill\null 
\columnbreak

 \begin{figure*}[t]
	\includegraphics[width=2\columnwidth]{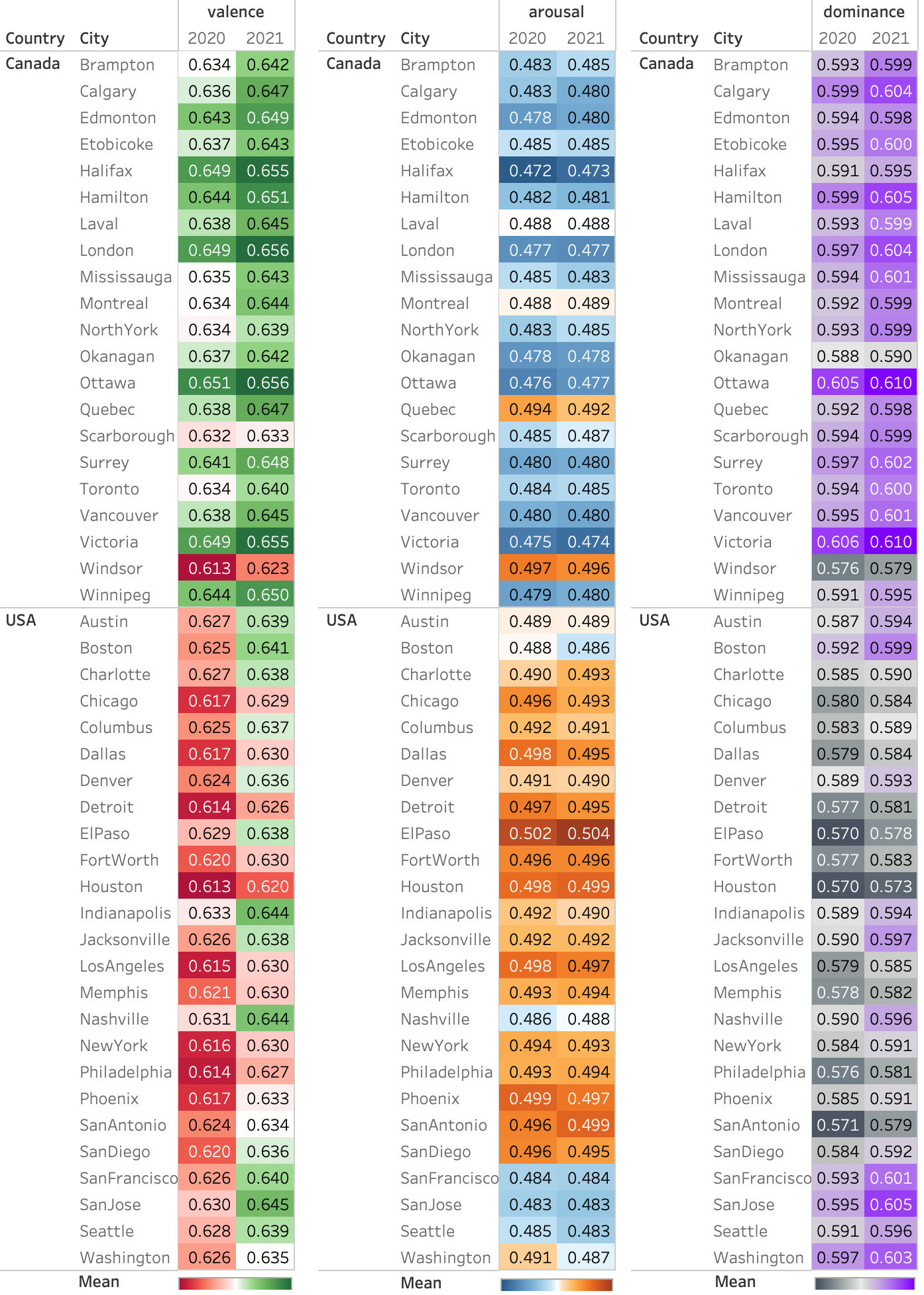}
 	\caption{TED: Tweet Valence Means (left), Arousal Means (centre), and Dominance Means (right) across American and Canadian cities in 2020 and 2021 (using tweets from TUSC-City).}
 	\label{fig:city-speaker-mean}
 \end{figure*}

 \begin{figure*}[t]
	\includegraphics[width=2\columnwidth]{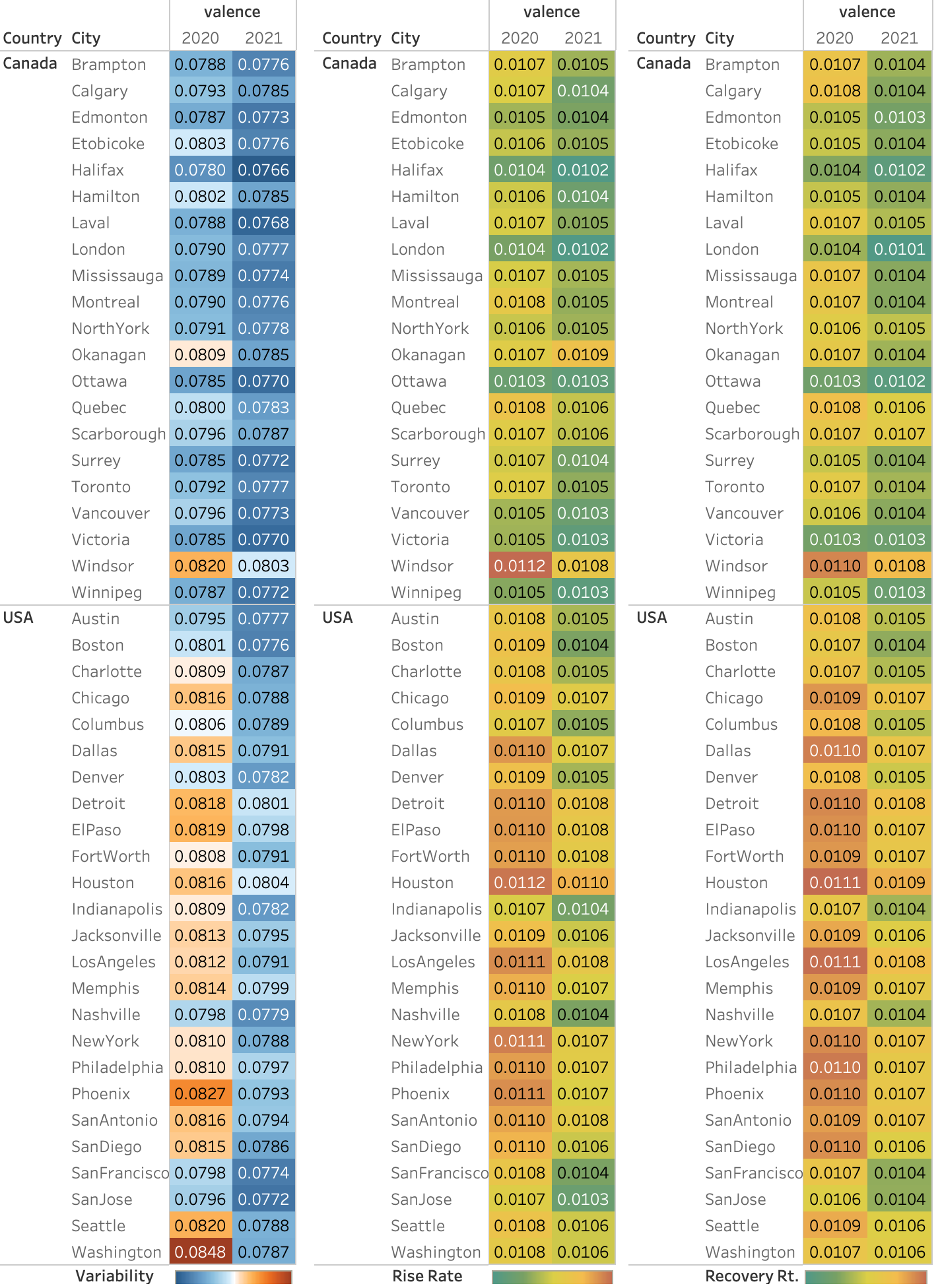}
 	\caption{TED: Tweet Valence Variability (left), Rise Rate (centre), and Recovery Rate (right) across American and Canadian cities in 2020 and 2021 (using tweets from TUSC-City).}
 	\label{fig:city-speaker-rest}
 \end{figure*}






\end{document}